\documentclass[10pt,twocolumn,letterpaper]{article}

\usepackage[pagenumbers]{wacv} 

\usepackage{graphicx}
\usepackage{amsmath}
\usepackage{amssymb}
\usepackage{booktabs}
\usepackage{algorithm}
\usepackage{algpseudocode}
\usepackage{multirow}
\usepackage{caption}

\usepackage[pagebackref,breaklinks,colorlinks]{hyperref}

\usepackage[capitalize]{cleveref}
\crefname{section}{Sec.}{Secs.}
\Crefname{section}{Section}{Sections}
\Crefname{table}{Table}{Tables}
\crefname{table}{Tab.}{Tabs.}

\def\wacvPaperID{2752} 
\def\confName{WACV}
\def\confYear{2025}

\begin{document}

\title{Universal Photorealistic Style Transfer: \\A Lightweight and Adaptive Framework}

\author{Rong Liu$^{1,2}$\;   Enyu Zhao$^1$\;   Zhiyuan Liu$^1$\;   Andrew Feng$^2$\;  Scott John Easley$^1$\\
$^1$University of Southern California\\ $^2$USC Institute for Creative Technologies\\
}
\maketitle


\begin{abstract}
Photorealistic style transfer aims to apply stylization while preserving the realism and structure of input content. However, existing methods often encounter challenges such as color tone distortions, dependency on pair-wise pre-training, inefficiency with high-resolution inputs, and the need for additional constraints in video style transfer tasks. To address these issues, we propose a Universal Photorealistic Style Transfer (UPST) framework that delivers accurate photorealistic style transfer on high-resolution images and videos without relying on pre-training. Our approach incorporates a lightweight StyleNet for per-instance transfer, ensuring color tone accuracy while supporting high-resolution inputs, maintaining rapid processing speeds, and eliminating the need for pretraining. To further enhance photorealism and efficiency, we introduce instance-adaptive optimization, which features an adaptive coefficient to prioritize content image realism and employs early stopping to accelerate network convergence. Additionally, UPST enables seamless video style transfer without additional constraints due to its strong non-color information preservation ability. Experimental results show that UPST consistently produces photorealistic outputs and significantly reduces GPU memory usage, making it an effective and universal solution for various photorealistic style transfer tasks.
\end{abstract}  
\section{Introduction}
\label{sec:intro}

Photorealistic style transfer aims to transform an image or video by applying a target style while preserving the structural and photorealistic integrity of the content. This technique has a wide range of practical applications, including virtual content creation, game development, video editing, and cinematic post-production. One of the key application areas is video style transfer, which allows color tones of scenes to be adjusted using reference style images. As the demand for ultra-high-definition content in modern cinema grows, achieving cost-effective video style transfer solutions while maintaining cinematic quality has become a crucial challenge for the industry. This approach might be the potential solution to reduce the need for costly reshoots and complex lighting setups. 
Another significant application is in 3D reconstruction. With advancements in techniques like Neural Radiance Fields (NeRF) and 3D Gaussian Splatting (3DGS)~\cite{mildenhall2020nerfrepresentingscenesneural, mueller2022instant, barron2023zipnerf, li2023neuralangelo, kerbl3Dgaussians, Huang2DGS2024, liu2024atomgs}, using style transfer to recreate scenes under different lighting conditions or appearances without additional data collection has become important. Furthermore, in multi-frame style transfer tasks, handling high-resolution frames or sequences introduces challenges related to computational and memory resources. Ensuring consistency across frames—whether temporal in videos or multi-view in 3D reconstruction—imposes another level of complexity. Many existing methods struggle to balance computational efficiency and consistency for large-scale, high-resolution inputs, underscoring the need for more scalable solutions for real-world photorealistic style transfer.

Early photorealistic style transfer methods \cite{li2017universal,li2018closed,yoo2019photorealistic} had an unsatisfying performance as their output demonstrated distortion and unwanted artifacts. Meanwhile, the processing speed and memory cost have severely limited their availability. Performing deterministic color mapping can be one possible solution for dealing with artifacts. The filter-based methods \cite{hu2018exposure,ke2022harmonizer} lack the ability to generalize complex color transformation due to filter limitations. 
In contrast, LUT-based methods \cite{xia2020joint,lin2023adacm} introduce another challenge: LUTs contain a large number of parameters, making them difficult to optimize. To circumvent this, some approaches propose alternatives to directly predicting LUTs. For example, Ke \etal \cite{ke2023neural} introduced a two-stage color mapping process that enables faster style transfer. While promising, these methods still deliver suboptimal results and rely on pair-wise datasets for training, which restricts their application to images and videos. 
When it comes to video style transfer, additional constraints are typically required to ensure temporal coherence and consistency across frames, complicating the process further \cite{ruder2018artistic,chen2017coherent,huang2017real,deng2021arbitrary,xia2021real}.

To this end, we propose Universal Photorealistic Style Transfer (UPST) which can perform style transfer on the high-resolution inputs in real-time, without any training dataset needed or extra constraints imposed, and produce comparative results, if not better, against state-of-the-art methods. Our contributions are as follows:

\begin{enumerate}

    \item We propose a lightweight StyleNet that performs per-instance style transfer without requiring pre-training on paired datasets. This neural network is designed to efficiently handle a wide range of input resolutions, from low to 8K, ensuring resolution universality while maintaining low memory overhead and high processing speeds.

    \item We introduce an instance-adaptive optimization strategy that uses an adaptive coefficient to adjust the balance between stylization and content preservation and adopts early stopping mechanisms to speed up convergence.

    \item Our Universal Photorealistic Style Transfer (UPST) framework naturally supports consistent multi-frame and video style transfer. By preserving non-color information and requiring no additional constraints, it provides a universal solution for both single images and sequential data.

\end{enumerate}

\begin{figure*}[htbp]
    \centering
    \includegraphics[width=0.8\linewidth]{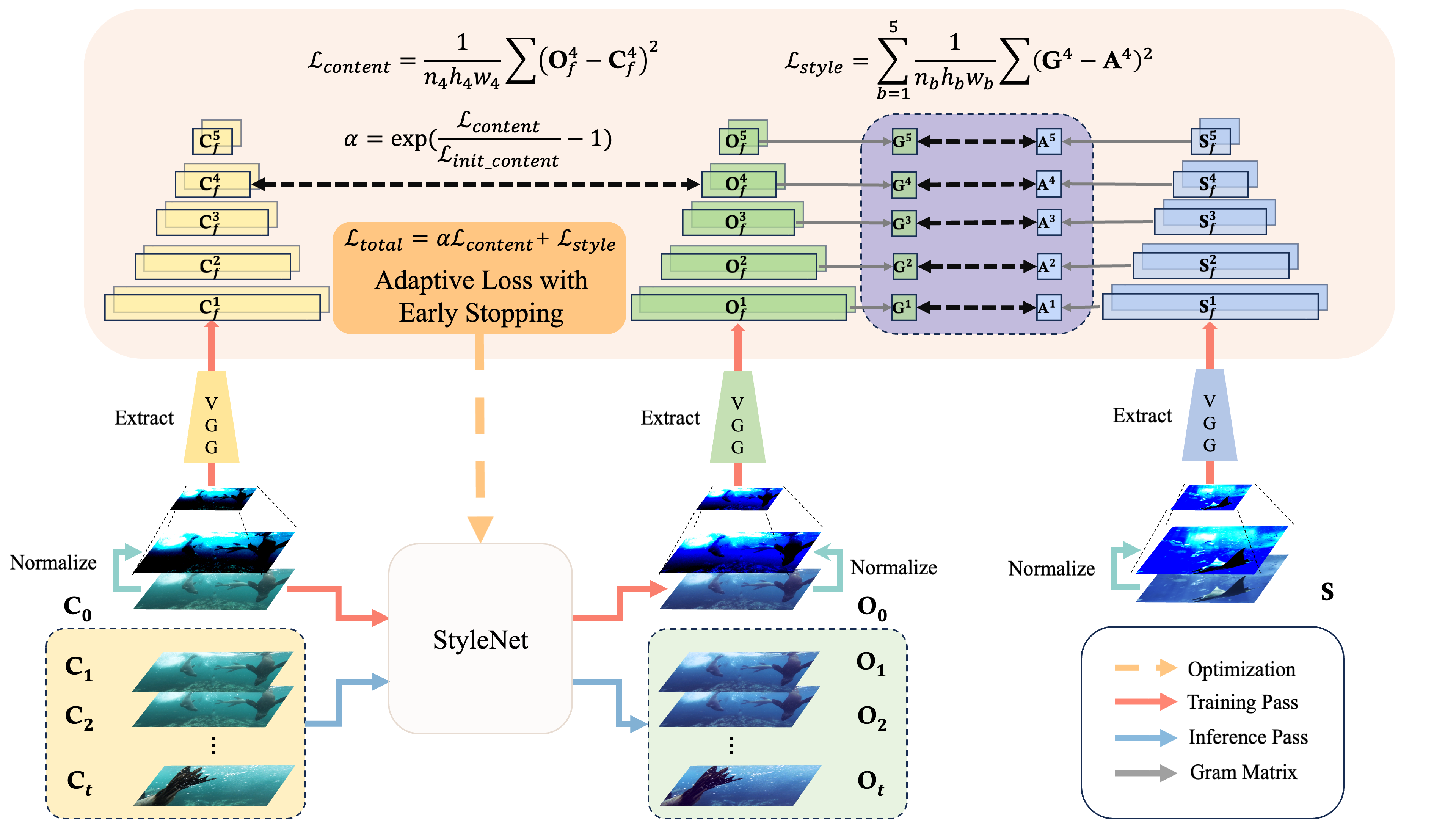}
    \caption{\textbf{Overview of Universal Photorealistic Style Transfer (UPST) Pipeline and Instance-Adaptive Optimization}. 
    }
    \label{fig:pipeline}
\end{figure*}

\section{Related Work}
\label{sec:relatedwork}

\subsection{Image Photorealistic Style Transfer} The advent of neural style transfer, introduced by Gatys \etal \cite{gatys2016image}, marked a significant shift towards using neural networks for style transfer. Artistic style transfer methods \cite{huang2023quantart,an2021artflow,huang2017arbitrary,li2019learning,deng2020arbitrary} have achieved remarkable results by transforming the style of real-world photos into artistic renditions. However, photorealistic style transfer presents different challenges, particularly the need to preserve the spatial structure of the input content after stylization, without introducing distortions or artifacts. While Gatys \etal \cite{gatys2016image} laid the foundation for style transfer, their results suffer from spatial distortions and unnatural color variations.

Subsequent approaches, such as WCT \cite{li2017universal} and PhotoWCT \cite{li2018closed}, reformulated style transfer as a whitening and coloring transform to improve photorealism. While these methods mitigate some issues, they still struggle to eliminate distortions and artifacts. Yoo \etal \cite{yoo2019photorealistic} advanced this line of work by using wavelet pooling/unpooling to better preserve high-frequency details, thus reducing distortion. Chiu \etal \cite{Chiu_2022_WACV} further improved image quality by introducing a block-wise coarse-to-fine training framework with high-frequency skip connections.

An alternative direction involves deterministic color mapping for style transfer. While filter-based methods \cite{hu2018exposure,ke2022harmonizer} maintain spatial structure, they often lack generalization, performing well only on specific image pairs and limited to simple adjustments like brightness or hue. More sophisticated approaches by Xia \etal \cite{xia2020joint} and Lin \etal \cite{lin2023adacm} leverage color look-up tables (LUTs) for deterministic color mapping. However, due to the high complexity of LUTs, direct optimization is infeasible. Instead, they simplify the process by using techniques such as a bilateral grid for LUT simplification \cite{xia2020joint} or predicting LUT parameters via a neural network \cite{lin2023adacm}. Ke \etal \cite{ke2023neural} proposed a two-stage method that first normalizes the content image's color and then applies deterministic color mapping to achieve the target style.

Despite these advancements, existing methods face three main limitations: the need for paired training datasets, difficulty in completely preventing distortions and artifacts, and high computational demands. In contrast, our method addresses these limitations by eliminating the need for paired datasets, minimizing computational overhead, and maintaining spatial fidelity without artifacts, while preserving non-color information.

\subsection{Multi-frame Photorealistic Style Transfer} While significant progress has been made in single-image style transfer, extending these techniques to multi-frame sequences introduces new challenges. Ensuring consistency across frames is crucial to avoid flickering and temporal artifacts. Existing video style transfer methods attempt to address this by enforcing temporal consistency through various constraints, such as temporal coherence loss between consecutive frames \cite{ruder2018artistic,chen2017coherent,xia2021real} or aligning style features with content features across the video \cite{deng2021arbitrary}. These approaches require additional loss functions or complex feature coupling mechanisms to maintain consistency, which can increase computational complexity and processing time.

Our approach, by contrast, naturally ensures temporal consistency across frames without the need for explicit constraints or losses. By focusing on preserving non-color information and leveraging instance-adaptive optimization, our framework avoids flickering and maintains coherence throughout the video, providing a more efficient and robust solution for multi-frame and video style transfer.

\section{Universal Photorealistic Style Transfer}
\label{sec:approach}

The Universal Photorealistic Style Transfer (UPST) framework, illustrated in \Cref{fig:pipeline}, introduces a novel approach to photorealistic style transfer that excels across diverse image resolutions, from low resolution to 8K, and ensures seamless transitions across both single image and sequential frames. UPST integrates three key components: StyleNet, which efficiently supports all resolution inputs; Instance-Adaptive Optimization, guaranteeing accurate color tone adjustments; and a two-stage process tailored to handle both static and dynamic data.

Traditional photorealistic style transfer approaches often rely on pre-training models with large paired content-style datasets, which are difficult to obtain due to the subjectivity of tone transfer and the infeasibility of guaranteeing accurate training pairs. UPST circumvents this limitation by reformulating style transfer as a per-instance optimization task, allowing precise control over stylization without compromising content structure, while giving users full flexibility over the target style.

To overcome the inherent inefficiency of per-instance optimization, UPST introduces a lightweight StyleNet combined with an Instance-Adaptive Optimization strategy. This integration accelerates the convergence process by dynamically adjusting the balance between stylization and content preservation, optimizing each instance independently and speeding up the process through early stopping mechanisms. As a result, UPST achieves fast yet accurate style transfer, meeting the demands of real-world applications.

Another major advantage of the UPST framework is its ability to maintain structural integrity across frames without the need for additional constraints. Once StyleNet is trained on the first frame of a video sequence, it can be directly applied to infer subsequent similar frames, eliminating the need for further fine-tuning and achieving speedups of up to 1000x over traditional methods. This capability makes UPST particularly suitable for high-resolution video style transfer tasks, offering both computational efficiency and temporal consistency.

In the following sections, we will explore each component of the UPST framework in detail. \Cref{sec:stylenet} focuses on the architecture of the lightweight StyleNet, while \Cref{sec:opt} elaborates on the instance-adaptive optimization method.

\subsection{Lightweight StyleNet}
\label{sec:stylenet}

\begin{figure*}[htbp]
    \centering
    \includegraphics[width=0.8\linewidth]{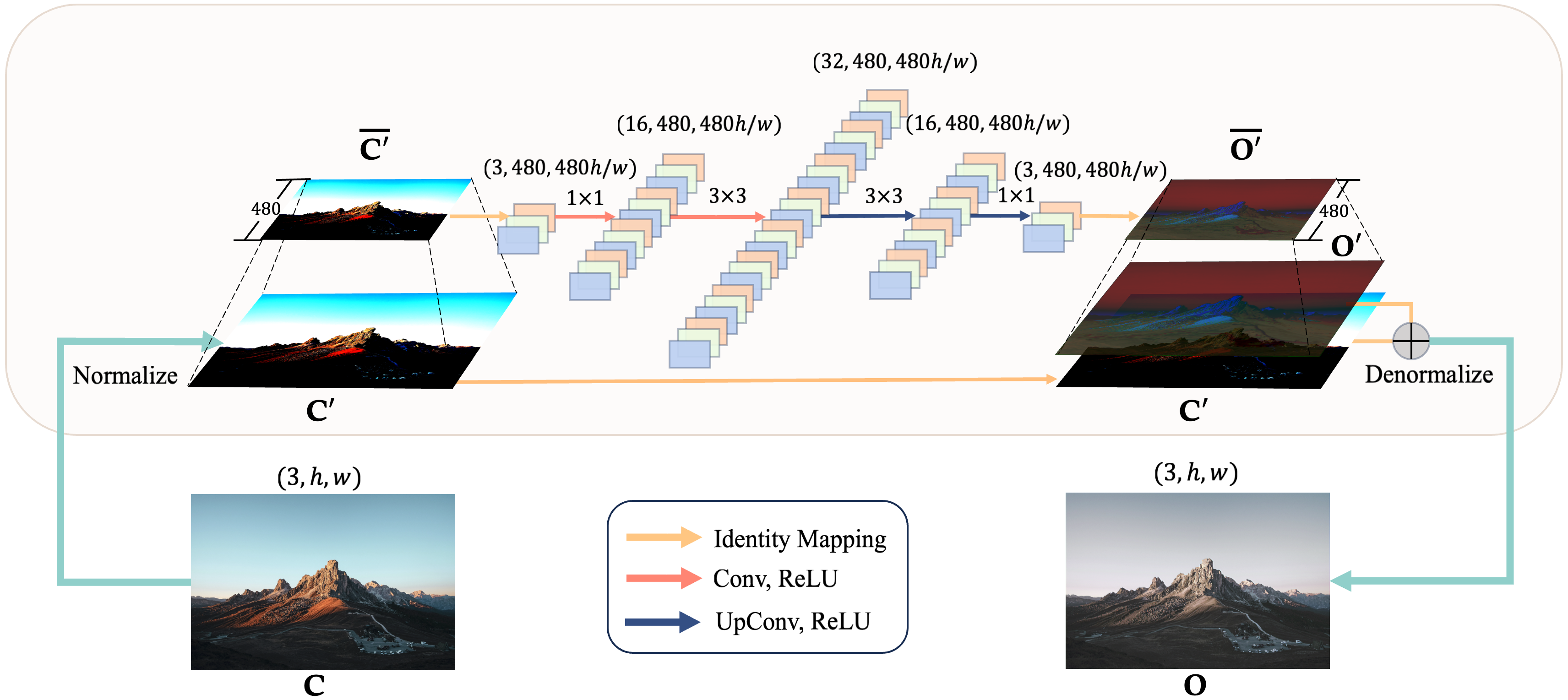}
    \caption{\textbf{Architecture of Lightweight StyleNet}. 
    }
    \label{fig:stylenet}
\end{figure*}

The architecture of StyleNet is depicted in \Cref{fig:stylenet}. StyleNet performs transformations within a normalized space using mean and standard deviation values derived from the ImageNet. 
Therefore, we begin by defining a normalization and denormalization function. The normalization function maps RGB color space to the normalized space $N:[0,1]^{3\times h\times w} \rightarrow \mathbb{R}^{3 \times h\times w}$. Given an RGB image $\mathbf{I}$, the mean $\mathbf{M}$ and standard deviation $\mathbf{\Sigma}$ of the ImageNet, the normalization function is defined as follows:

\begin{equation}
    N(\mathbf{I}) = (\mathbf{I} - \mathbf{M}) \oslash \mathbf{\Sigma},
\end{equation}
where $\oslash$ denotes the element-wise division. 
Conversely, to convert the normalized output $\mathbf{I}^\prime$ back to the RGB space for the final result, a denormalization function is utilized $D: \mathbb{R}^{3\times h\times w} \rightarrow [0,1]^{3\times h\times w}$, which is given by:
\begin{equation}
    D(\mathbf{I}^\prime) = \mathbf{I}^\prime \odot \mathbf{\Sigma} + \mathbf{M},
\end{equation}
where $\odot$ represents the element-wise multiplication.

Next, we illustrate the procedure of StyleNet. Given an input content image $\mathbf{C} \in \mathbb{R}^{3\times h\times w}$, StyleNet first normalizes it by
\begin{equation}
    \mathbf{C}^\prime = N(\mathbf{C}).
\end{equation}
Then, it diverges into two branches: a style transfer branch and a content shortcut branch. 

In the style transfer branch, the normalized image $\mathbf{C}^\prime$ is initially downsampled to Standard Definition (SD) resolution (480p) to facilitate resolution-agnostic and efficient transformation computations:
\begin{equation}
    \overline{\mathbf{C}^\prime} = \downarrow \mathbf{C}^\prime, \quad \overline{\mathbf{C}^\prime} \in \mathbb{R}^{3 \times 480\times \frac{480h}{w}},
\end{equation}
where $\downarrow$ is the downsampling operation.

Subsequently, we formulate the color transformation in the style transfer branch as a segmentation task where the targeted output is the whole color space $[0,1]^3$ instead of being restricted to specific classification labels. Hence, the transformation is defined as $f: \mathbb{R}^{3 \times 480\times \frac{480h}{w}} \rightarrow \mathbb{R}^{3 \times 480\times \frac{480h}{w}}$. Here, $f$ is a Convolutional Neural Network (CNN) inspired by U-Net \cite{ronneberger2015u}. 
It follows a symmetrical structure, with the first half comprising Convolutional layers and the second half consisting of UpConvolutional layers. The initial part encodes the input into feature maps within a latent space, progressively expanding the number of channels. The subsequent part decodes these feature maps into a segmentation mask, serving as a color transformation mask that is applied to the input content.

\begin{table}[htbp]
    \centering
    
    \begin{tabular}{cccc}\hline
        Kernel & Padding & Input Channel & Output Channel \\ \hline
        1$\times$1 & 0 & 3 & 16\\
        3$\times$3 & 1 & 16 & 32\\
        3$\times$3 & 1 & 32 & 16\\
        1$\times$1 & 0 & 16 & 3\\ \hline
    \end{tabular}
    \caption{Hyperparameter of the transformation $f$}
    \label{tab:stylenet}
\end{table}

Notably, we adopt a proper hyperparameter setting, as detailed in \Cref{tab:stylenet}, to ensure the preservation of spatial information during the transformation performed by $f$. This table also facilitates the calculation of trainable parameters in the transformation $f$, totaling $9,379$. This lightweight design enables the potential for instant photorealistic style transfer applications.
With the definition of the transformation $f$ and the downsampled image $\overline{\mathbf{C}^\prime}$, we generate color transformation mask output $\overline{\mathbf{O}^\prime}$ by 
\begin{equation}
    \overline{\mathbf{O}^\prime} = f(\overline{\mathbf{C}^\prime}).
\end{equation}

Following the relatively intensive transformation computation, we upsample the mask back to the original resolution $\mathbf{O}^\prime$ by
\begin{equation}
 \mathbf{O}^\prime = \uparrow \overline{\mathbf{O}^\prime}, \quad
\mathbf{O}^\prime \in \mathbb{R}^{3\times h\times w},   
\end{equation}
where $\uparrow$ denotes the upsampling operation.

Simultaneously with the color transformation mapping, an identity mapping establishes a shortcut in StyleNet, aiding in preserving non-color information in the final output.

By combining these two branches, we generate the ultimate output $\mathbf{O}$ of StyleNet:
\begin{equation}
    \mathbf{O} = D(\mathbf{C}^\prime + \mathbf{O}^\prime).
\end{equation}

\subsection{Instance-Adaptive Optimization}
\label{sec:opt}

In order to enable photorealistic priority and rapid convergence of StyleNet, we develop an instance-adaptive optimization method, drawing inspiration from deep image representations introduced by Gatys \etal \cite{gatys2016image}, to train StyleNet.

Given an image denoted as $\mathbf{I}$, the feature map extracted by VGG \cite{simonyan2014very} in the first layer of the CNN block $b$ is represented as $\mathbf{I}^b_f\in \mathbb{R}^{n_b \times h_bw_b}$, where $n_b$ stands for the channel number of the feature map, and $h_b$ and $w_b$ are the height and width of the feature map.

With this feature map definition, given the content image $\mathbf{C}$ and output result $\mathbf{O}$, we can compute the feature maps of the fourth block, namely $\mathbf{C}^4_f$ and $\mathbf{O}^4_f$. The content loss can then be defined as follows:

\begin{equation}
    \mathcal{L}_{content}(\mathbf{C}, \mathbf{O})= \frac{1}{n_4h_4w_4}\sum(\mathbf{C}^4_f-\mathbf{O}^4_f)^2.
\end{equation}

Moreover, the style loss is assessed using Gram matrices between two feature maps. Given a feature map $\mathbf{I}^b_f$, the Gram matrix is computed as follows:
\begin{equation}
    \mathbf{G}^b = \mathbf{I}^b_f \mathbf{I}^{b\top}_f,\quad\mathbf{G}^b \in \mathbb{R}^{n_b \times n_b}.
\end{equation}

Assuming that $\mathbf{G}^b$ and $\mathbf{A}^b$ are the style Gram matrices for the output result $\mathbf{O}$ and style image $\mathbf{S}$ in block $b$, the style loss is defined as:
\begin{equation}
    \mathcal{L}_{style}(\mathbf{O}, \mathbf{S})= \sum_{b=1}^5\frac{1}{n_bh_bw_b}\sum(\mathbf{G}^b-\mathbf{A}^b)^2.
    \label{eq:style_loss}
\end{equation}

With the definition of content and style loss, the total loss is expressed as:
\begin{equation}
    \mathcal{L}_{total}(\mathbf{C}, \mathbf{O}, \mathbf{S})= \alpha\mathcal{L}_{content}(\mathbf{C}, \mathbf{O}) + \mathcal{L}_{style}(\mathbf{O}, \mathbf{S}),
\end{equation}
where $\alpha$ serves as a hyperparameter that controls the weight ratio between content and style. However, $\alpha$ often requires manual tuning and remains fixed during the entire training process, which can hinder photorealistic quality, especially when content and style losses evolve at different rates.

\begin{algorithm}
\caption{Instance-Adaptive Optimization}\label{alg:adaptive_optimization}
\begin{algorithmic}
\renewcommand{\algorithmicrequire}{\textbf{Input:}}
 \renewcommand{\algorithmicensure}{\textbf{Output:}}
\Require $\mathbf{C},\mathbf{S},\text{StyleNet}$
\Ensure $\text{StyleNet}$
\State $\text{optimizer}\gets\text{Adam(lr=0.001)}$
\State $\text{best\_loss} \gets \infty$
\State $\text{patience} \gets 10$
\While{$\text{patience} \neq 0 $}

\State $\mathbf{O} \gets \text{StyleNet}(\mathbf{C})$ 
\State $\text{content\_loss, style\_loss} \gets \text{get\_loss}(\mathbf{C}, \mathbf{O}, \mathbf{S})$
    \If{$\text{best\_loss}=\infty$} \Comment{First epoch}
        \State $\text{initial\_content\_loss}\gets\text{content\_loss}$
        \State $\text{initial\_total\_loss}\gets\text{content\_loss}+\text{style\_loss}$
    \EndIf
    \State $\alpha \gets \text{exp}(\frac{\text{content\_loss}}{\text{inital\_content\_loss}}-1)$
    \State $\text{total\_loss}\gets \alpha \times\text{content\_loss}+\text{style\_loss}$
    \If{$\frac{\text{total\_loss}}{\text{initial\_total\_loss}}<\text{best\_loss}-0.01$}
        \State $\text{best\_loss} \gets \frac{\text{total\_loss}}{\text{initial\_total\_loss}}$
        \State $\text{patience} \gets 10$
    \Else
        \State $\text{patience} \gets \text{patience}-1$
    \EndIf
    \State $\text{StyleNet}\gets\text{optimizer(total\_loss)}$
\EndWhile
\\
\Return $\text{StyleNet}$
\end{algorithmic}
\end{algorithm}

To overcome these limitations, we propose an instance-adaptive optimization framework (Algorithm \ref{alg:adaptive_optimization}) where $\alpha$ dynamically adjusts based on content variation. Rather than treating $\alpha$ as a static scalar, we redefine it as an adaptive coefficient that evolves throughout the training process. Initially, we set $\alpha=1$, giving equal weight to both content preservation and style transfer objectives. As content loss begins to rise during stylization, $\alpha$ increases exponentially, shifting focus towards preserving content structure as style features are transferred. This adaptive mechanism improves both convergence stability and photorealism, reducing conflicts between content preservation and stylization.

Additionally, we implement an early stopping mechanism to expedite convergence and prevent overfitting. We monitor the change in the total loss relative to its initial value. If the total loss reduction falls below a $1\%$ threshold over ten consecutive epochs, training is halted early. This technique not only accelerates the training process but also reduces unnecessary computational costs, making the optimization more efficient overall.

\section{Experiments}
\label{sec:experiments}

\begingroup
\setlength{\tabcolsep}{1pt}
\begin{figure*}[htbp]
\centering
\begin{tabular}{ccccc}
  \includegraphics[width=.10\linewidth]{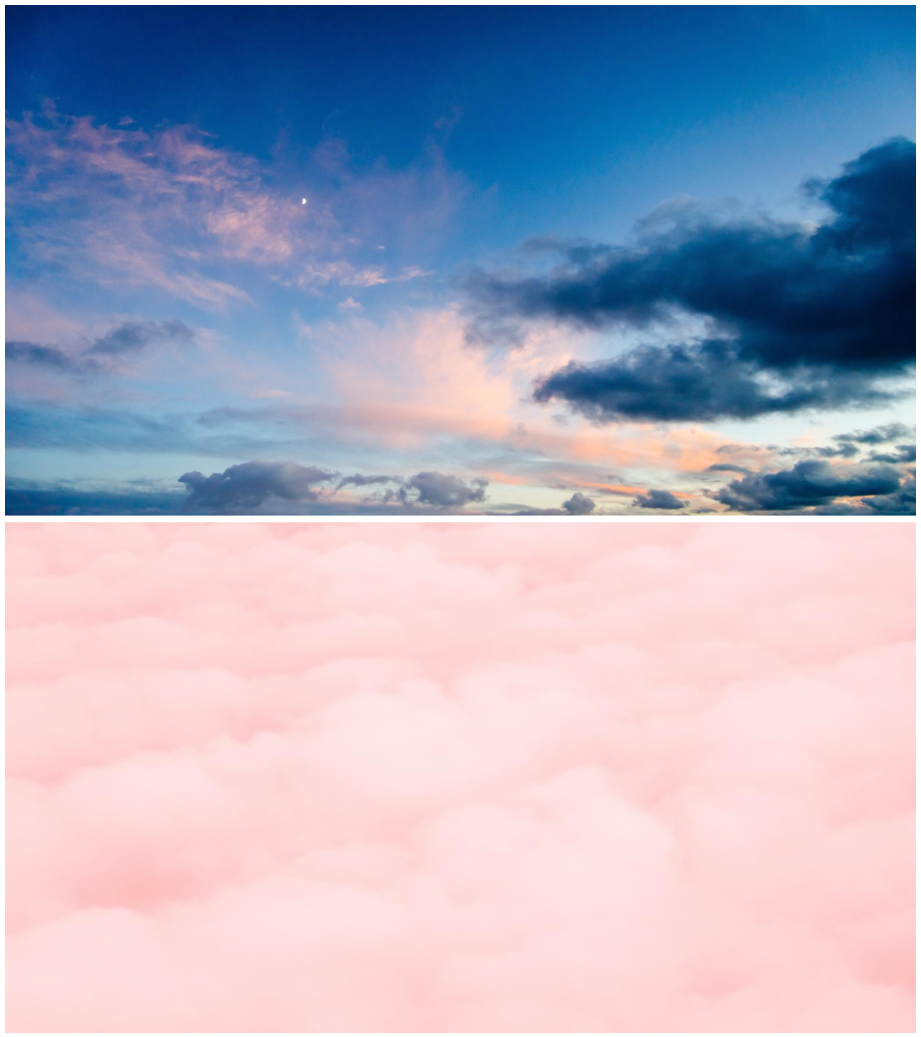} &    \includegraphics[width=.20\linewidth]{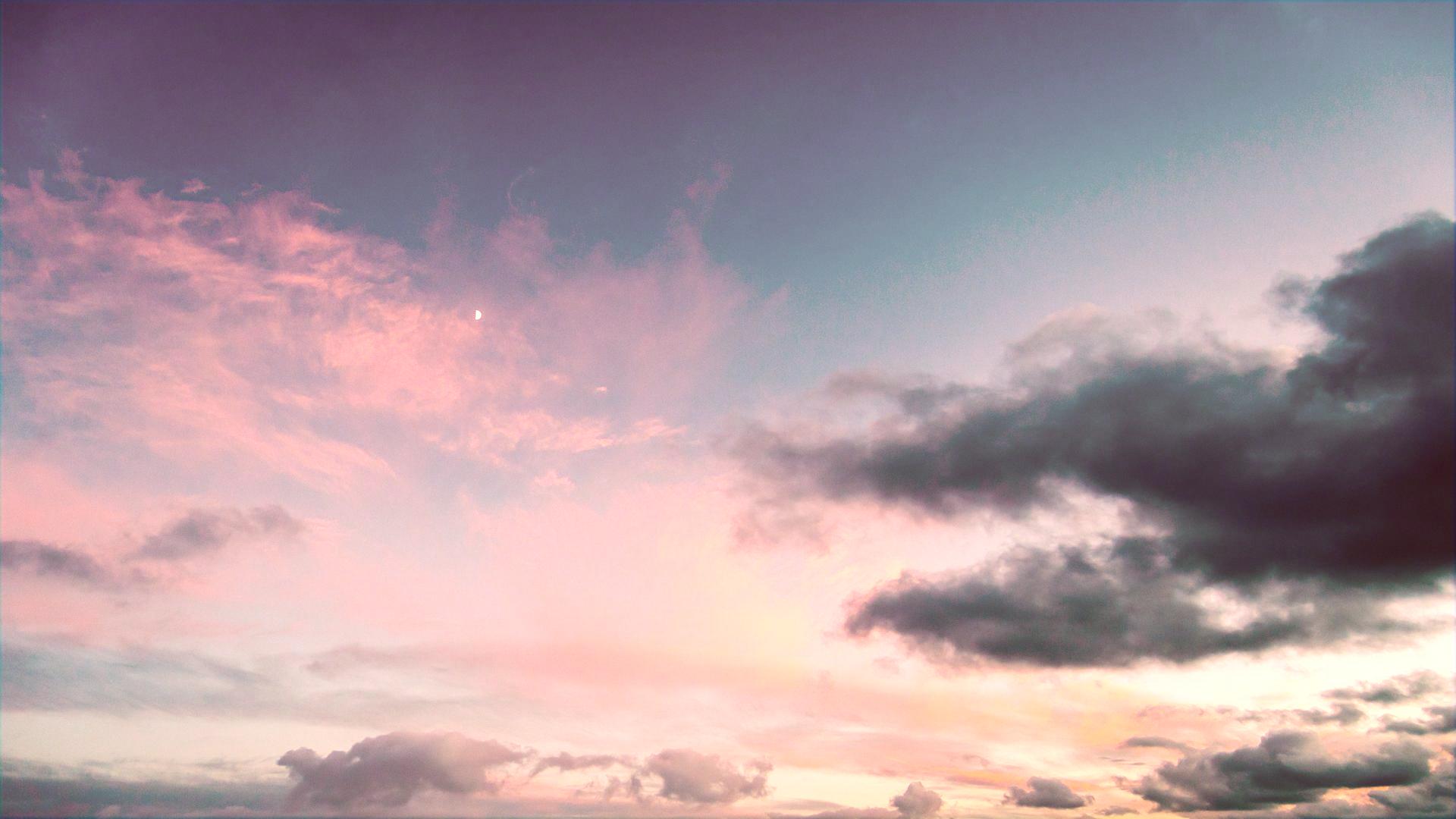}&   \includegraphics[width=.20\linewidth]{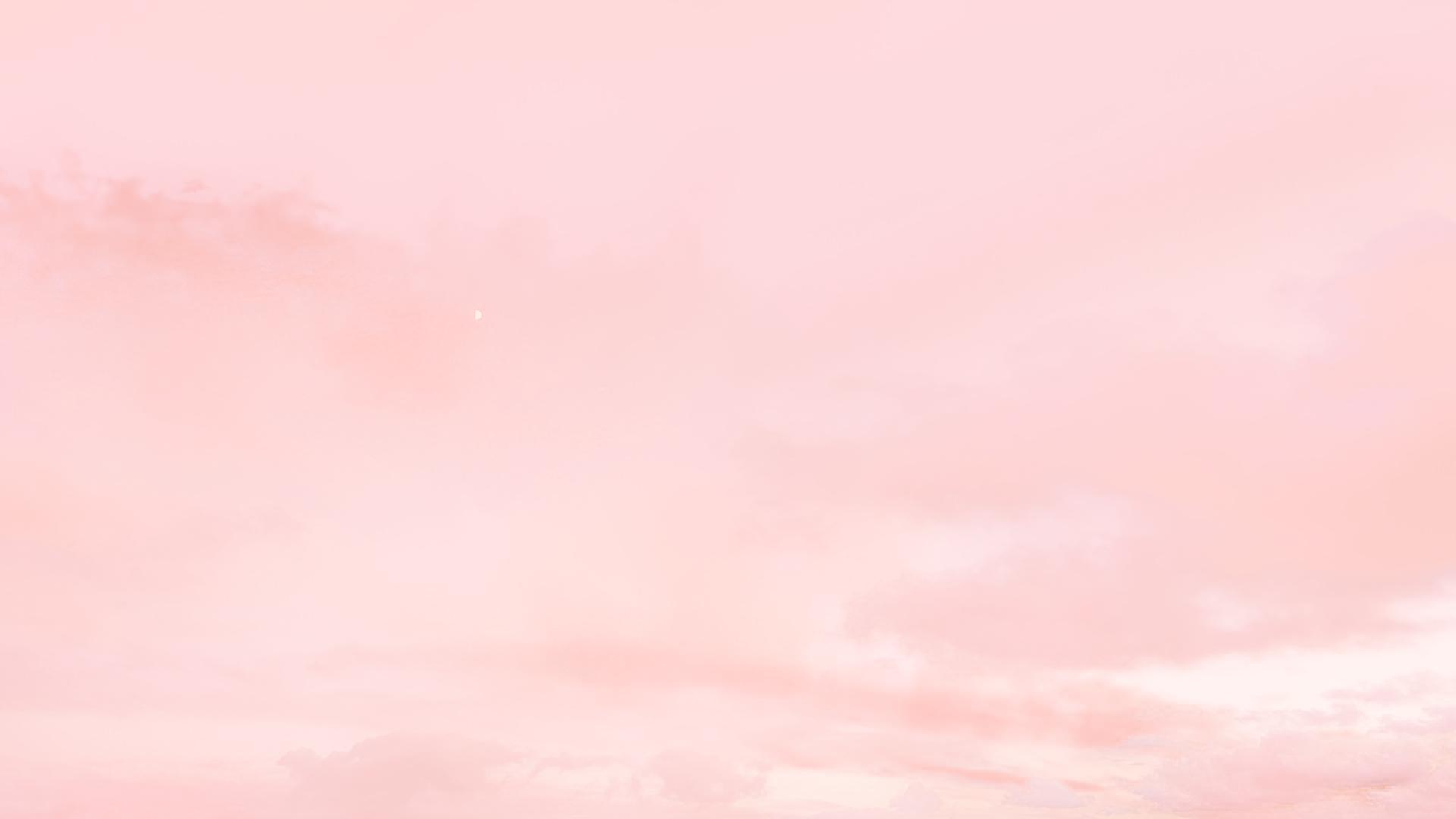}&   \includegraphics[width=.20\linewidth]{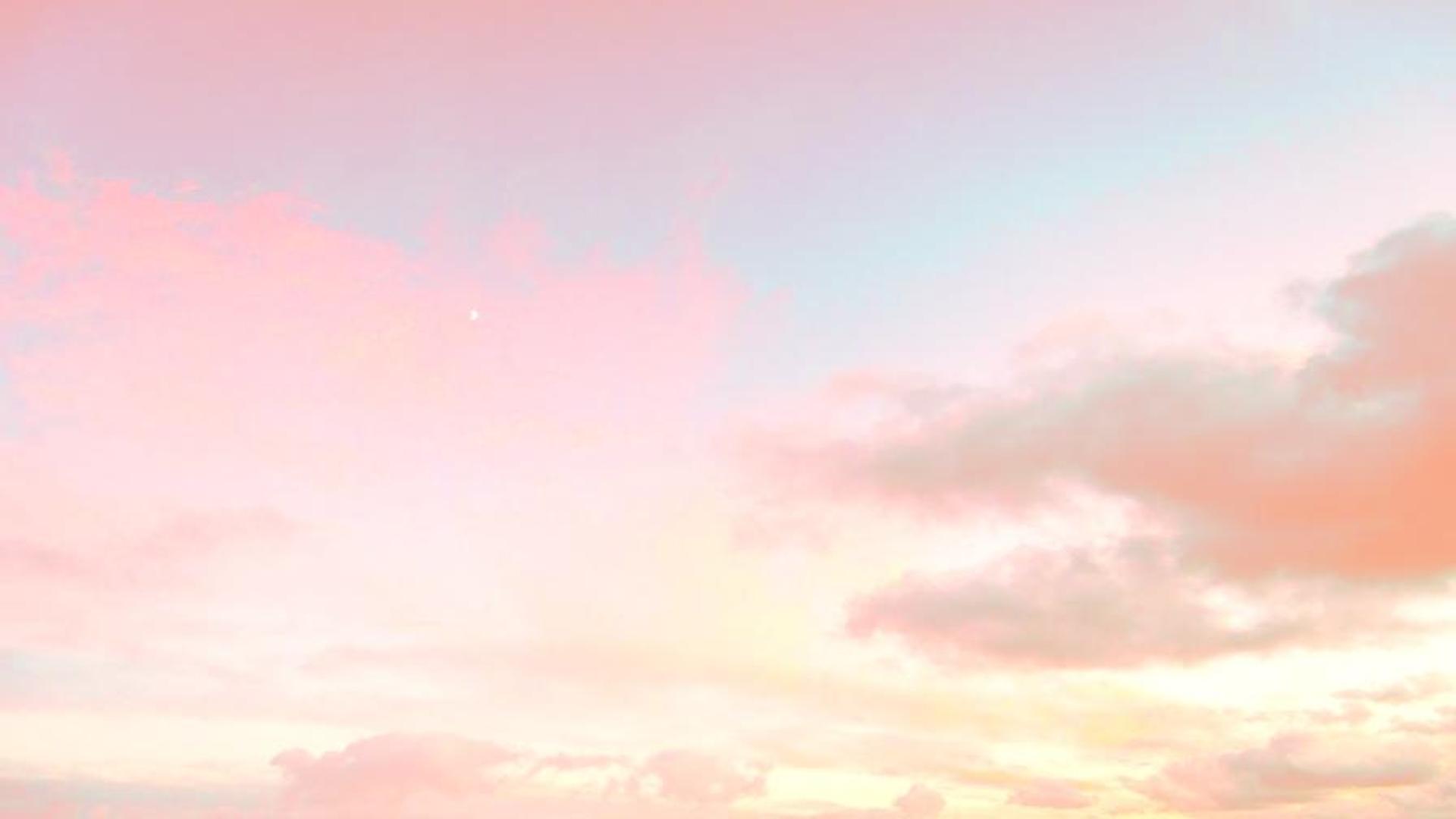}&   \includegraphics[width=.20\linewidth]{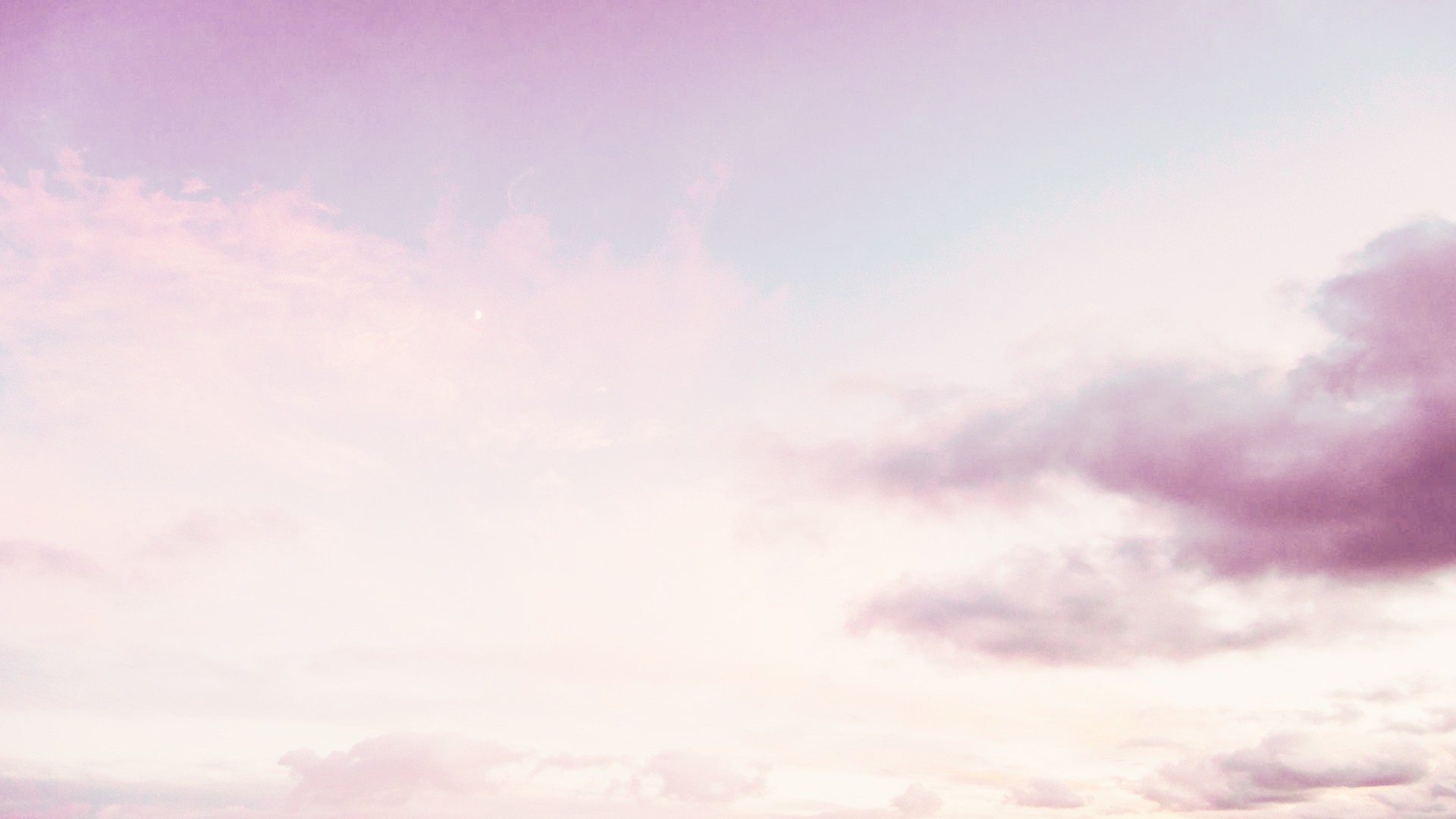}\\

  \includegraphics[width=.10\linewidth]{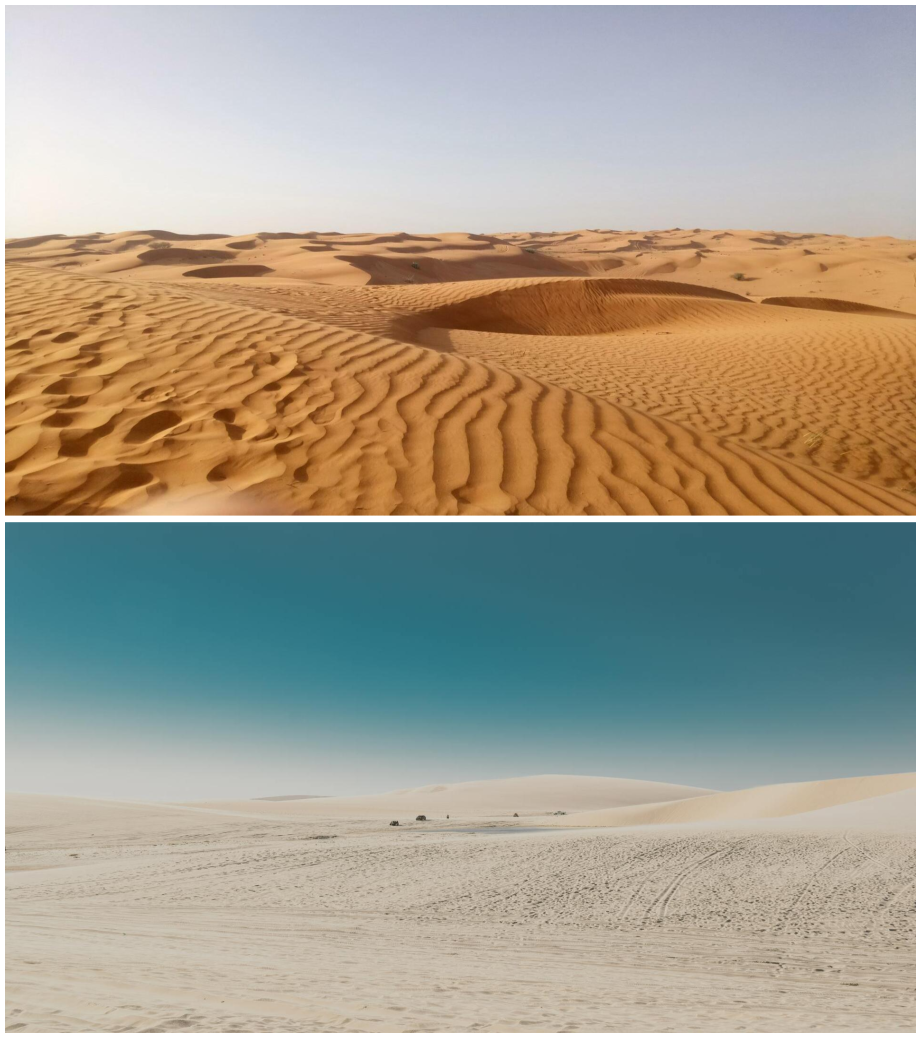} &     \includegraphics[width=.20\linewidth]{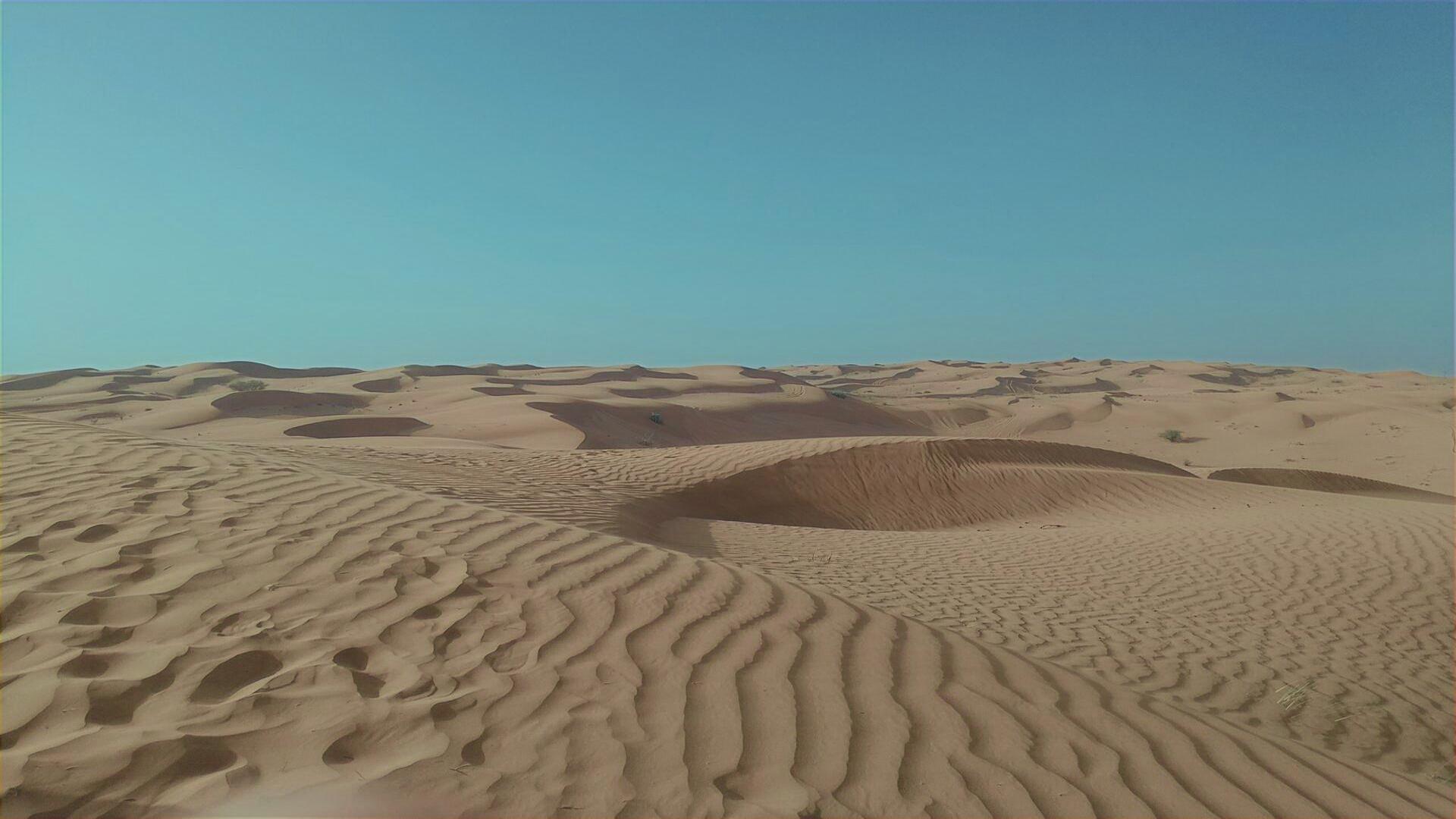}&   \includegraphics[width=.20\linewidth]{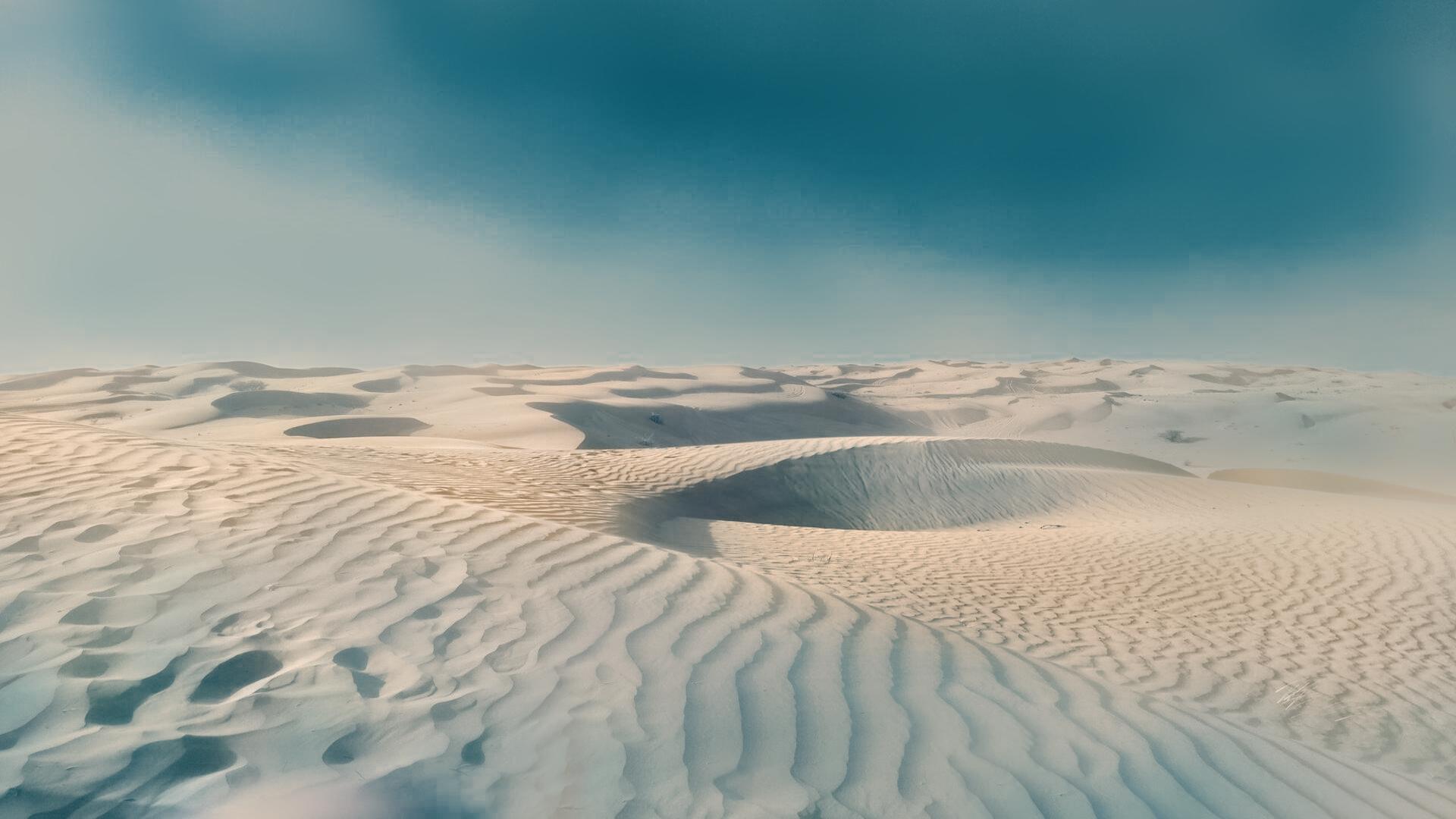}&   \includegraphics[width=.20\linewidth]{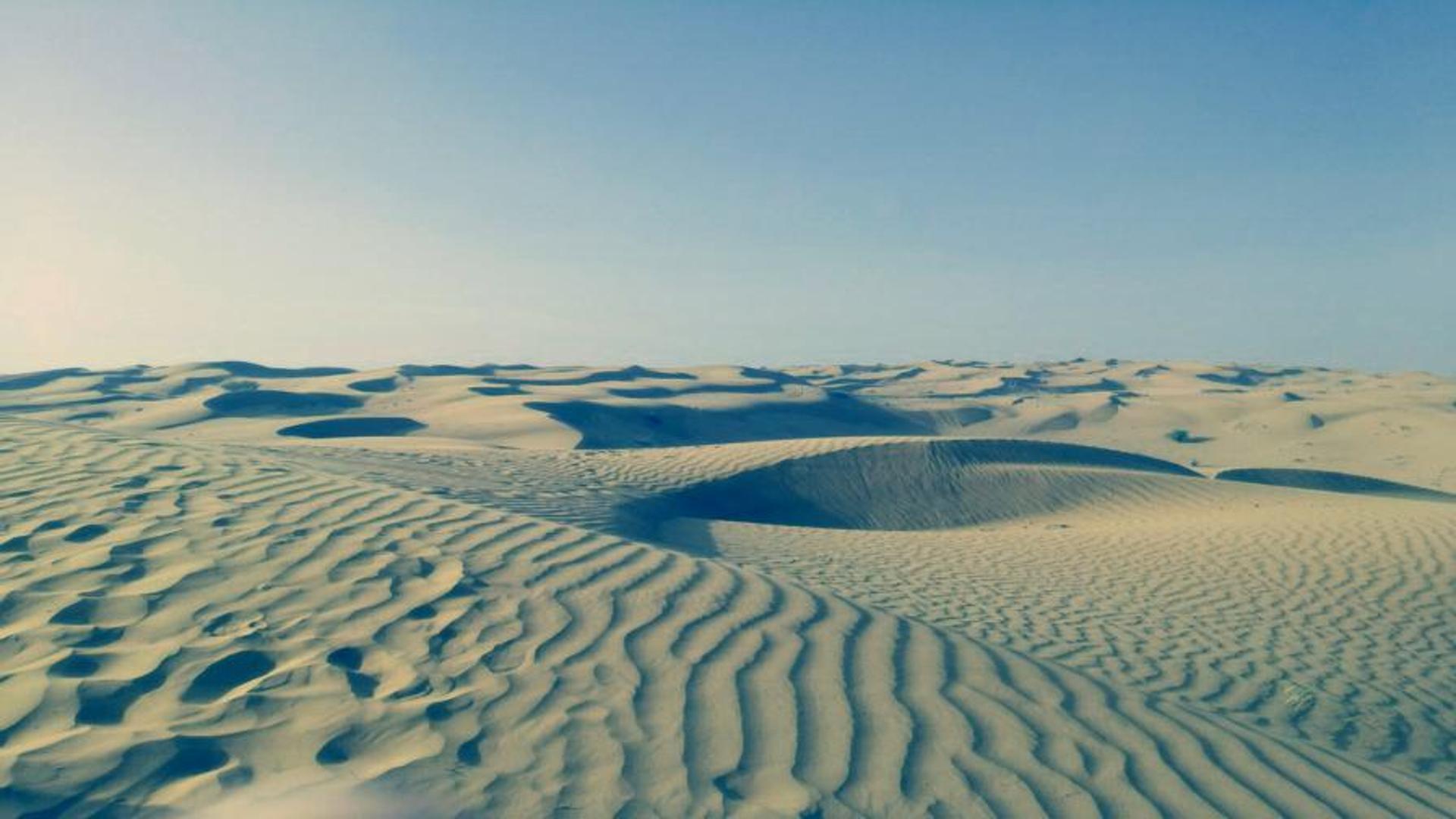}&   \includegraphics[width=.20\linewidth]{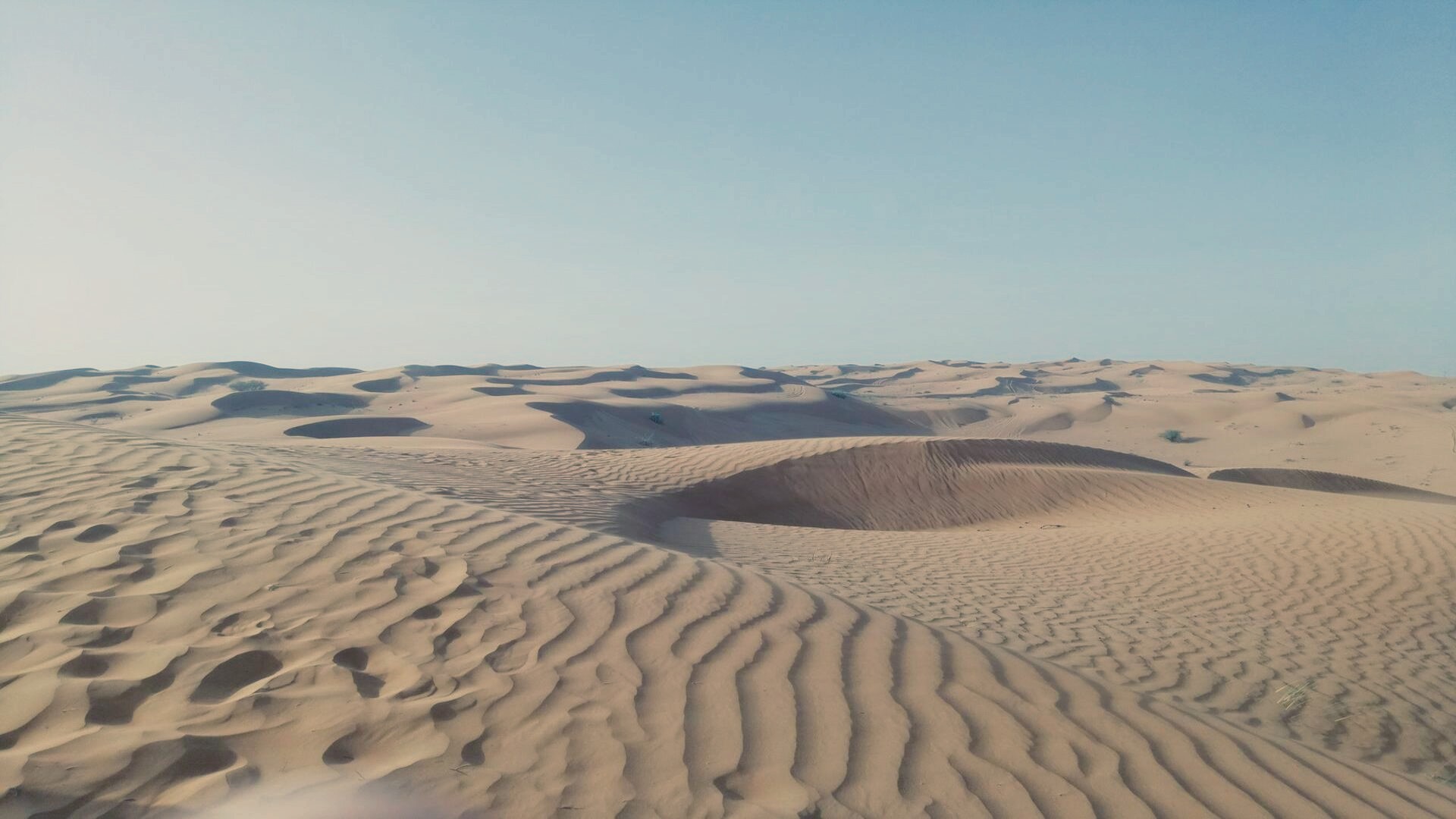} \\

  \includegraphics[width=.10\linewidth]{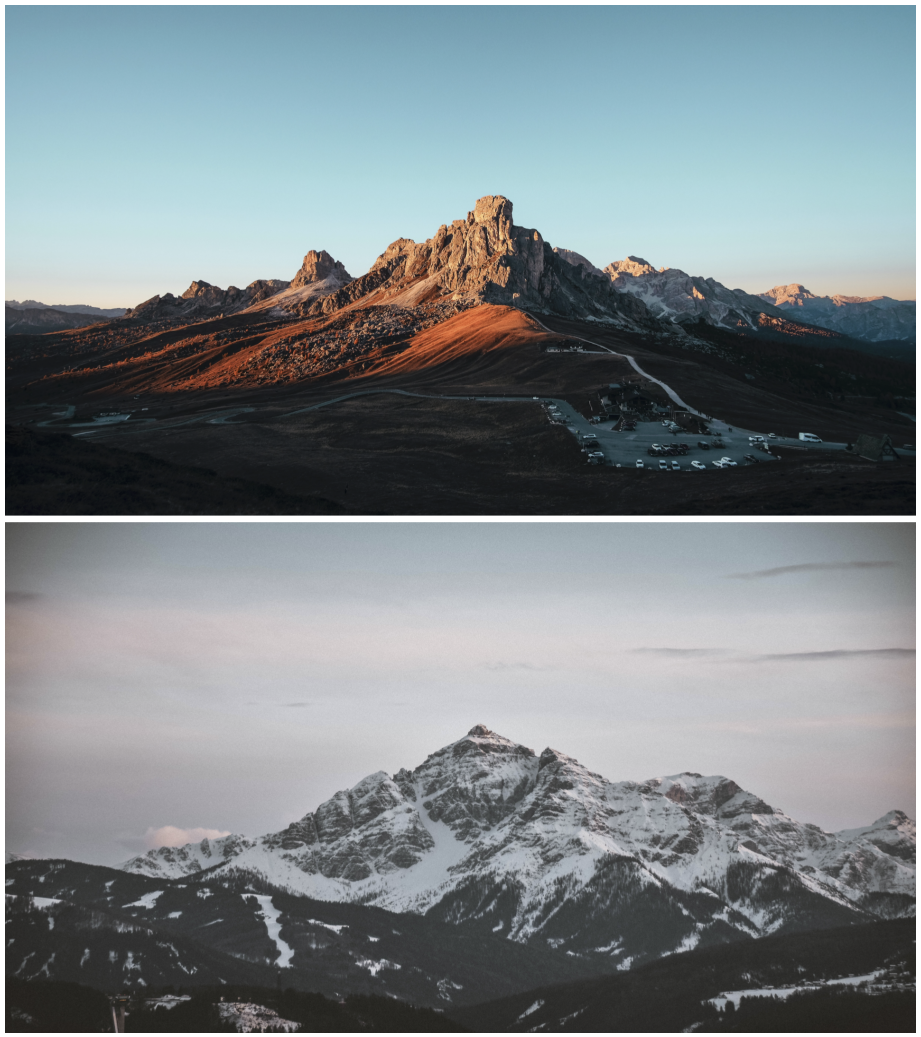} &     \includegraphics[width=.20\linewidth]{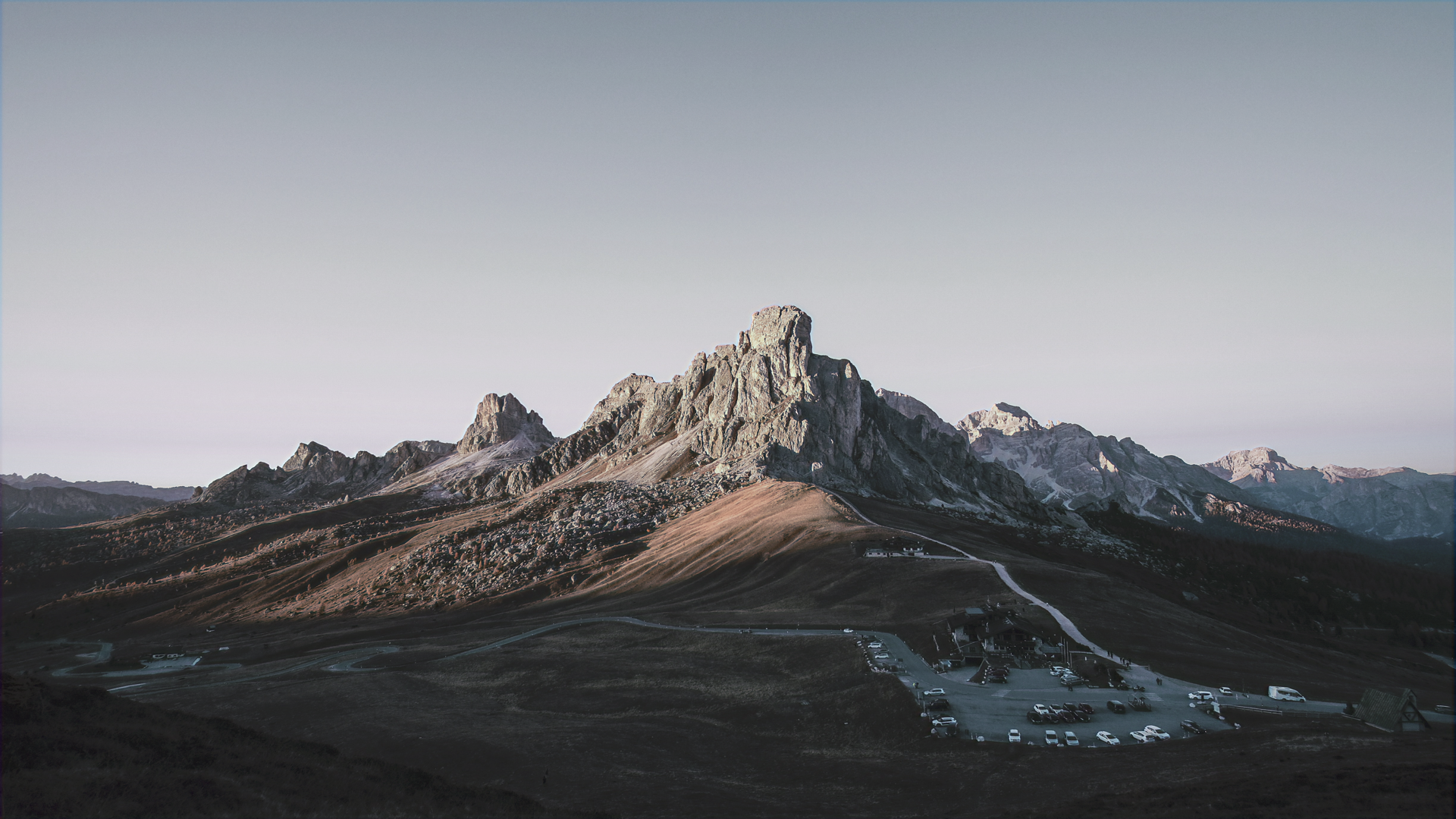}&   \includegraphics[width=.20\linewidth]{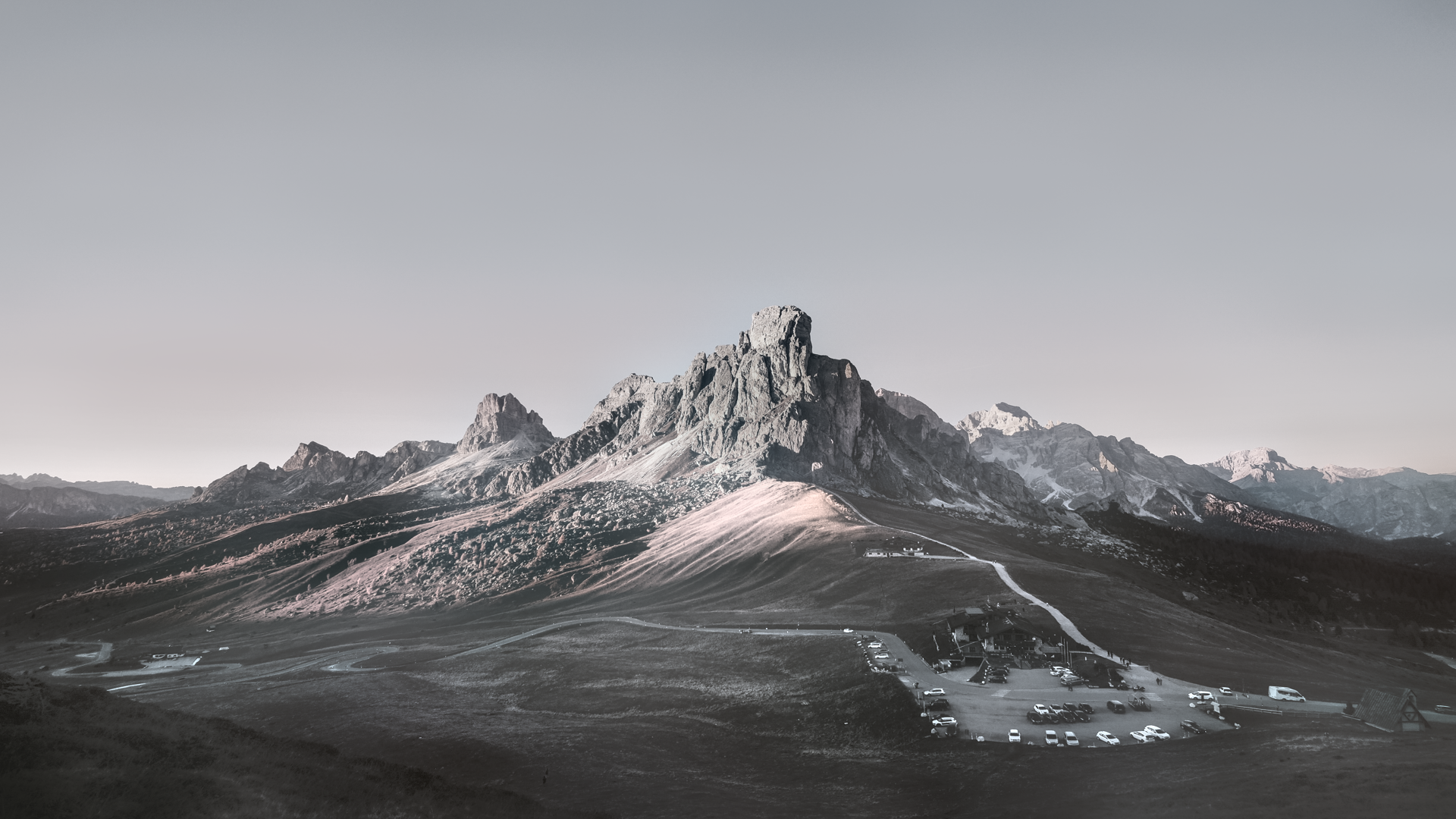}&   \includegraphics[width=.20\linewidth]{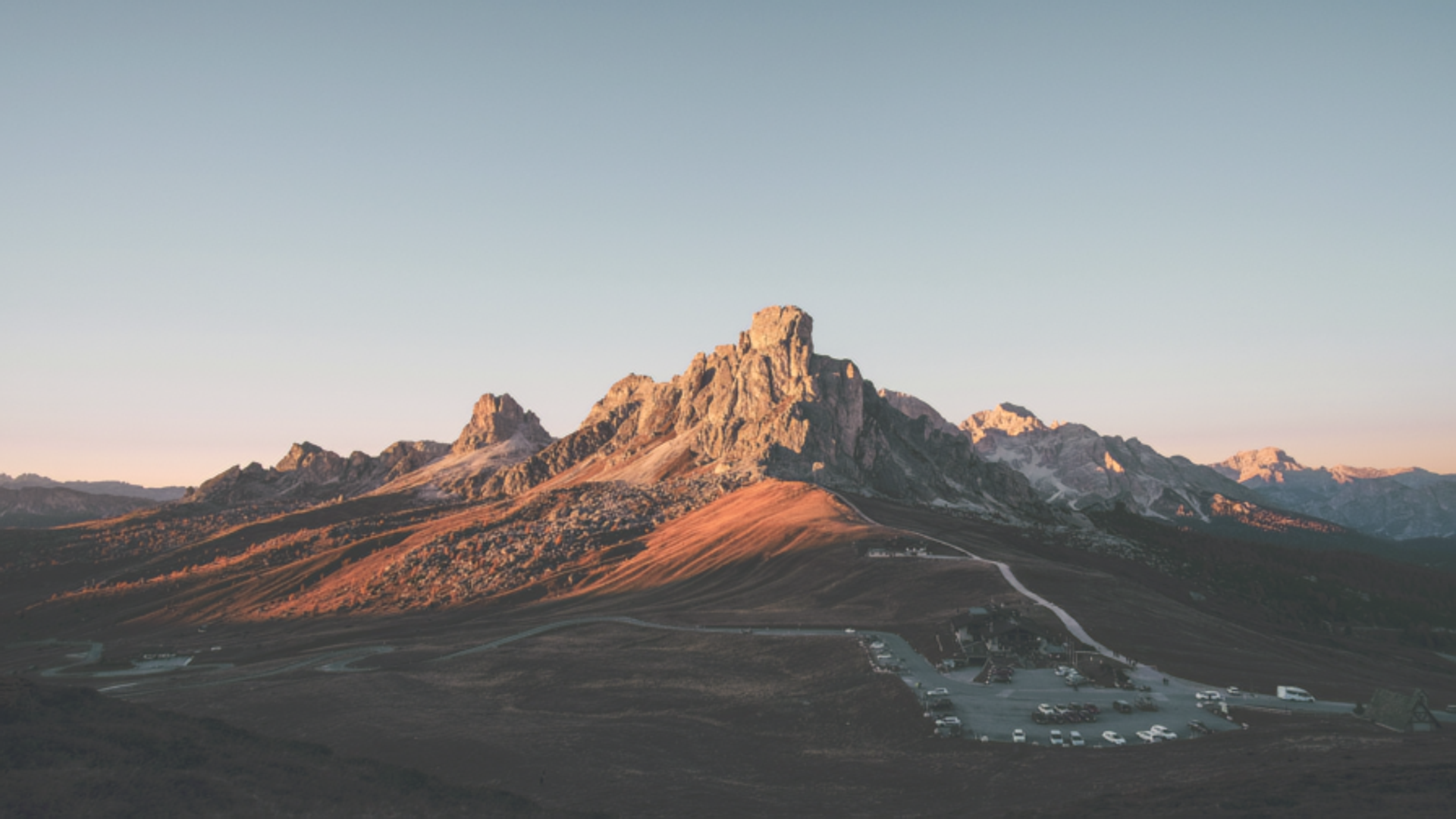}&   \includegraphics[width=.20\linewidth]{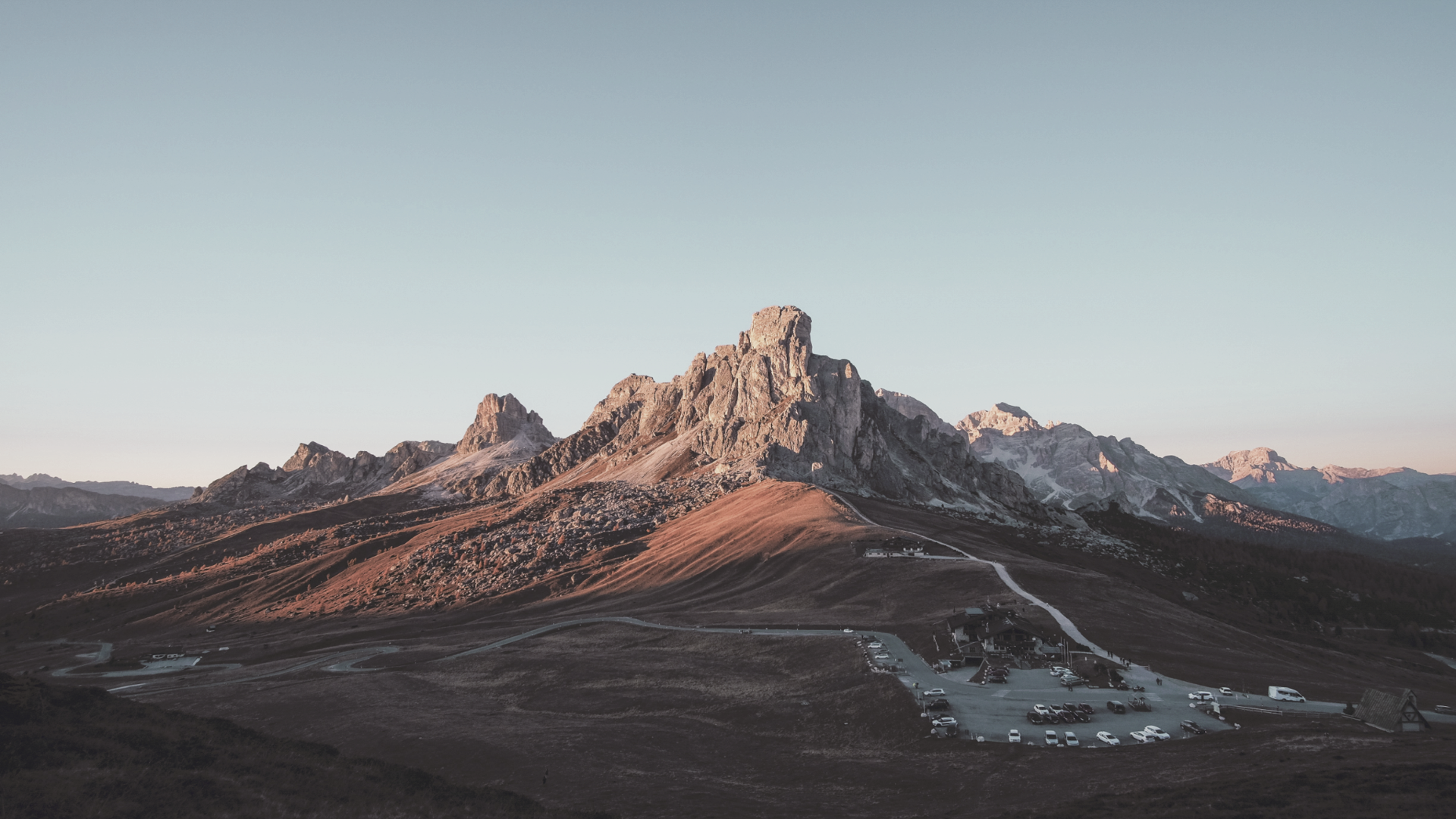} \\
  (a) Input &(b) Ours&(c) $\text{PhotoWCT}^2$ \cite{Chiu_2022_WACV}& (d) $\text{WCT}^2$ \cite{yoo2019photorealistic} &(e) Neural Preset \cite{ke2023neural}
\end{tabular}
\caption{\textbf{Qualitative Comparison.} Given (a) an input pair consisting of content (top) and style (bottom), the results produced by (b) Ours, (c) $\text{PhotoWCT}^2$ \cite{Chiu_2022_WACV}, (d) $\text{WCT}^2$ \cite{yoo2019photorealistic}, and (e) Neural Preset \cite{ke2023neural} are presented. Our approach outperforms others in prioritizing content photorealism and transferring desired style effects. Note that IPSI achieves these results without pre-training on annotated datasets.}
\label{fig:qualitative}
\end{figure*}
\endgroup

\begin{table*}[htbp]
\centering
\begin{tabular}{c|ccccc}
\hline
\multirow{2}{*}{Method} & \multicolumn{4}{c}{GPU Time$\downarrow$ / Memory Consumption$\downarrow$}                                                                                                                                                                                                      & Model Size$\downarrow$                                                     \\ \cline{2-6} 
                        & \begin{tabular}[c]{@{}c@{}}FHD\\ (1920 × 1080)\end{tabular} & \begin{tabular}[c]{@{}c@{}}2K\\ (2560 × 1440)\end{tabular} & \begin{tabular}[c]{@{}c@{}}4K\\ (3840 × 2160)\end{tabular} & \begin{tabular}[c]{@{}c@{}}8K\\ (7680 × 4320)\end{tabular} & \begin{tabular}[c]{@{}c@{}}Number of\\ Parameters\end{tabular} \\ \hline
$\text{PhotoWCT}^2$ \cite{Chiu_2022_WACV}                & 0.51 s / 13.47 GB                                              & 0.91 s / 17.47 GB                                             & 1.89 s / 23.36 GB                                               & OOM                                                        & 7.05 M                                                         \\
$\text{WCT}^2$ \cite{yoo2019photorealistic}                    & 0.99 s / 21.03 GB                                             & OOM                                                        & OOM                                                          & OOM                                                        & 10.12 M                                                        \\
Ours (Train)                  &    2.66 s / 3.22 GB                                                          &      2.75 s / 3.32 GB                                                      &    3.15 s / 3.62 GB                                                          &      6.34 s / 4.91 GB                                                      & 9379                                                           \\
Ours (Inference)                   &   0.74 ms / 0.67 GB                                                          &       0.91 ms / 0.75 GB                                                     &        1.38 ms / 0.95 GB                                                      &          3.63 ms / 2.06 GB                                                  & 9379       
                \\ \hline
\end{tabular}
    \caption{\textbf{Transfer Efficiency Comparison.} 
 The evaluation is conducted on an NVIDIA GeForce RTX 4090 GPU (24GB memory). The units “s”, “ms”, “GB”, and “M” mean seconds, milliseconds, gigabytes, and millions, respectively. “OOM” means out-of-memory issue.}
    \label{tab:transfer_efficiency}
\end{table*}

In this section, we first provide an overview of our experimental setup, involving details about the evaluation dataset and the quantitative metrics employed. Subsequently, we extensively compare UPST with other competing methods based on transfer effects and efficiencies. Furthermore, we delve into ablation studied on the StyleNet architecture and the Instance-adaptive optimization approach.

\noindent\textbf{Evaluation Dataset.} To conduct both quantitative and qualitative comparisons, we utilized the DPST dataset \cite{luan2017deep}, comprising sixty content-style pairs. This dataset encompasses a mix of pairs, some align roughly correct segmentation while others pose challenges. 

\begin{table}[htbp]
    \centering
    \begin{tabular}{cccc}\hline
         &  Content$\uparrow$&Style$\uparrow$\\ \hline
         Neural Preset \cite{ke2023neural}&0.7115$\pm$0.1542&0.7786$\pm$0.2434\\
         $\text{WCT}^2$ \cite{yoo2019photorealistic}&0.6172$\pm$0.1566&0.8166$\pm$0.2379\\
         $\text{PhotoWCT}^2$ \cite{Chiu_2022_WACV}&0.7007$\pm$0.1320&0.8234$\pm$0.2540\\
         Ours&\textbf{0.7484$\pm$0.1309}&\textbf{0.8568$\pm$0.2051}\\\hline
    \end{tabular}
    \caption{\textbf{Quantitative Comparison on DPST Dataset.}}
    \label{tab:quantitative_comparison}
\end{table}

\noindent\textbf{Quantitative Metrics.} Prior studies \cite{an2020ultrafast, xia2020joint, yoo2019photorealistic} have employed content structural similarity and Gram matrix style loss for quantitative evaluation of photorealistic style transfer results. Furthermore, Neural Preset \cite{ke2023neural} upgraded the edge detection model from HED \cite{xie2015holistically} to LDC \cite{soria2022ldc} to enhance the precision of structural similarity measurement. Building upon these previous works, we adopt LDC to assess the content similarity and utilize the normalized Gram matrix style loss to evaluate the style similarity. The style similarity is defined as $\text{Style} = \max(0,3000-\mathcal{L}_{style})/3000$,
where $3000$ is a relatively large style loss for normalization purposes. In an ideal scenario, the method exhibits high similarities in both content and style. Therefore, we also calculate the F-1 score that represents the harmonic mean of content and style similarities to offer a comprehensive evaluation. The F-1 score is defined as $F_1 = 2/(\text{Content}^{-1}+\text{Style}^{-1})$

\subsection{Comparisons}
We qualitatively and quantitatively compare UPST with $\text{WCT}^2$ \cite{yoo2019photorealistic}, $\text{PhotoWCT}^2$ \cite{Chiu_2022_WACV}, and Neural Preset \cite{ke2023neural}, for which the authors have released access to codes or demos.

\noindent\textbf{Qualitative Comparison.}
In \Cref{fig:qualitative}, we present a qualitative comparison highlighting the superior quality of our approach. 
Our approach applies an accurate photorealistic transformation that preserves high-quality non-color information, such as intricate textures and consistent brightness and contrast properties inherited from the content image (\eg the sky color tone, cloud texture, and contrast properties in \Cref{fig:qualitative} (b)). 
In contrast, $\text{PhotoWCT}^2$ overfits the style image and yields results closely resembling the style image, producing strong style-effect results with artifacts (\eg the sky artifacts and distortions in \Cref{fig:qualitative} (c)). On the other hand, $\text{WCT}^2$ blends the color tones from the style image, resulting in unatural color transfer (\eg the blue dessert in \Cref{fig:qualitative} (d)).
Neural Preset performs well by maintaining content structure from the content image. However, due to its pre-trained nature, it may not always accurately extract color information from the style image, occasionally resulting in color tone deviations of outputs (\eg the purple color tone in \Cref{fig:qualitative} (e)).

\begin{figure*}[htbp]
    \centering
    \includegraphics[width=0.9\linewidth]{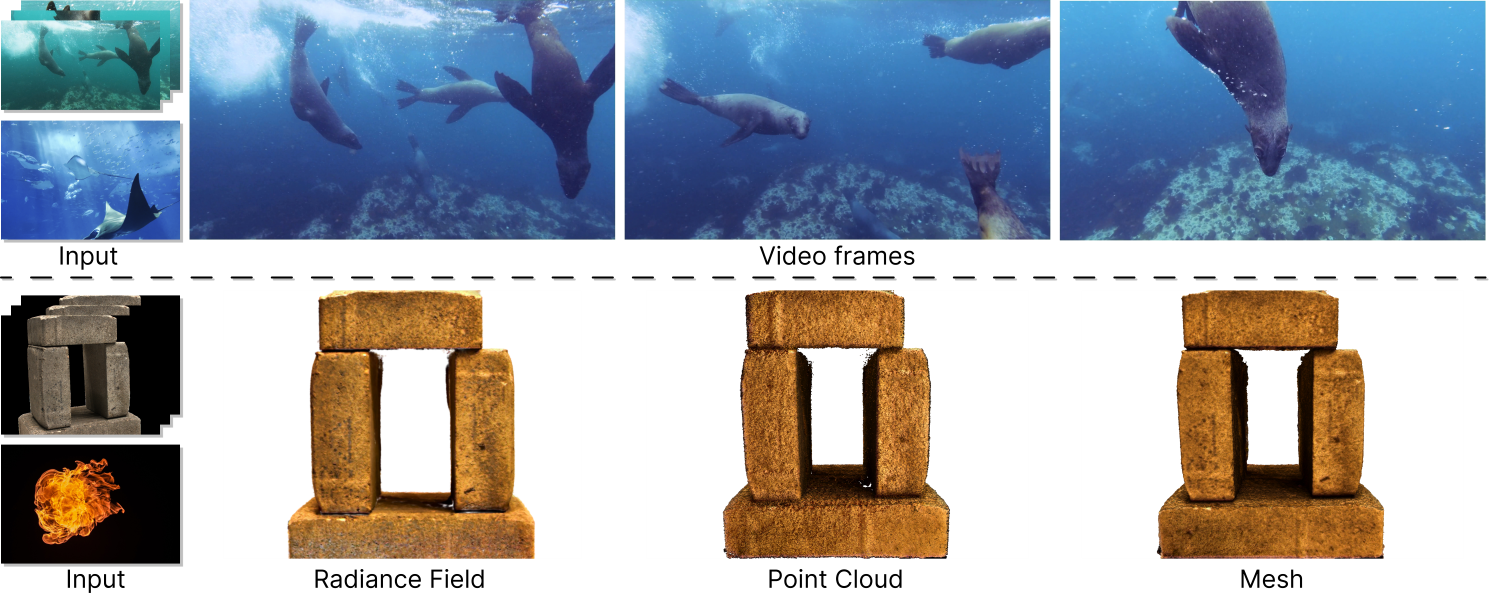}
\caption{\textbf{Multi-Frame Applications.}}
\label{fig:application}
\end{figure*}

\noindent\textbf{Quantitative Comparison.} 
The findings in \Cref{tab:quantitative_comparison} align consistently with those presented in \Cref{fig:qualitative}, affirming that our approach yields an overall superior performance with minimal standard deviation. In contrast, $\text{WCT}^2$ struggles to preserve the original resolution of content input so it fails to generate clear, high-quality results, resulting in lower overall scores. While $\text{PhotoWCT}^2$ excels in Style similarity, it suffers from overfitting issues, impacting its F-1 score due to low content similarity. On the other hand, the Neural Preset demonstrates a high content similarity but faces challenges in achieving a universal color transformation, resulting in a lower F-1 score.

\noindent\textbf{User Study.} To assess the subjective quality of various methods, we conducted a comprehensive user study involving 50 participants. Each participant was presented with 10 image sets randomly selected from our evaluation set. Every image set included an input content image, a reference style image, and 4 randomly shuffled color style transfer results. Participants were asked to rank the 4 results based on several criteria: content preservation, stylization accuracy, photorealism, and the visual appeal of the transferred results. The top 1 preferred rate and average ranking for each method were then computed to evaluate their performance, as shown in \Cref{tab:user_study}. Our method was largely preferred, achieving a top-1 selection rate of 48.4\%, which is significantly higher than PhotoWCT$^2$ (21.8\%), Neural Preset (19.4\%), and WCT$^2$ (10.4\%). Additionally, our method achieved the best average ranking (1.94), further indicating its superior performance across the evaluated criteria.

\begin{table}[htbp] 
\centering 
\begin{tabular}{ccccc} 
\hline & Top 1 Rate$\uparrow$ & Avg. Ranking$\downarrow$ \\ \hline
Neural Preset & 19.4\% & 2.25 \\ 
WCT$^2$ & 10.4\% & 2.95 \\ 
PhotoWCT$^2$ & 21.8\% & 2.86 \\ 
 Ours & \textbf{48.4\%} & \textbf{1.94} \\ 
\hline 
\end{tabular} 
\caption{\textbf{User Study Comparison.}} \label{tab:user_study} 
\end{table}

\noindent\textbf{Transfer Efficiency.} 
Neural Preset \cite{ke2023neural} is not included in this transfer efficiency comparison, as its code has not been publicly available.
As presented in \Cref{tab:transfer_efficiency}, $\text{PhotoWCT}^2$ \cite{Chiu_2022_WACV} and $\text{WCT}^2$ \cite{yoo2019photorealistic} demand significant memory resources for the transfer process and are unable to handle high-resolution transfer tasks. In contrast, our method only requires a small memory and excels at handling high-resolution transfer tasks. To ensure the universal transfer accuracy, our method requires per-instance optimization during the transfer stage, which results in slower but comparable transfer speeds compared to others if the input is a single frame. However, our method outperforms them when the input is multi-framed.
In our approach, after using the first frame to train the StyleNet, the subsequent frames will be inferred by trained StyleNet, leading to approximately 1000 times transfer efficiency. 


\subsection{Multi-Frame Applications}
\label{sec:application}

UPST excels in handling multi-frame transfer tasks due to its ability to preserve non-color information, including temporal and multi-view consistency from the input, coupled with its remarkable amortized transfer efficiency.
In \Cref{fig:application}, we showcase two distinct multi-frame applications: Video Style Transfer (top row) and 3D Appearance Transfer (bottom row). For the 3D reconstruction task, we first apply style transfer to the calibrated image set, which is then fed into AtomGS~\cite{liu2024atomgs} to generate different 3D representations such as a radiance field, colored point cloud, and textured mesh. The results show that the transferred video frames maintain high temporal consistency, while the successful reconstruction confirms the multi-view consistency of the style transfer.

However, our approach assumes that the input is a short video clip to prevent drift commonly encountered in longer videos.
In scenarios involving longer videos with drastic scene changes, UPST can still be applied by selecting key frames from long videos and providing corresponding style images for those frames. 
This strategy allows UPST to handle extended videos while maintaining high-quality style transfer results, effectively managing scene transitions without sacrificing temporal or stylistic consistency.

\begin{table*}[htbp]
    \centering
    \resizebox{0.8\linewidth}{!}{
    \begin{tabular}{cccccc}\hline
         &  Best setting&w/o adaptive optimization&4x channels&0.25x channels&w/o spatial preservation\\ \hline
         Content& 0.7484&0.6460&0.6834&\textbf{0.7872}&0.7001\\
         Style& 0.8568&\textbf{0.9075}&0.8637&0.7656&0.7398\\
         F-1 & \textbf{0.7989}&0.7548&0.7630&0.7762&0.7194\\
         Speed (fps)&\textbf{0.4464}&0.2532&0.4444&0.3367&0.3921\\ \hline
    \end{tabular}
    }
    \caption{\textbf{Quantitative Comparison of Ablation Studies.}}
    \label{tab:ablation_table}
\end{table*}

\begin{figure*}[htbp]
    \centering
    \includegraphics[width=0.98\linewidth]{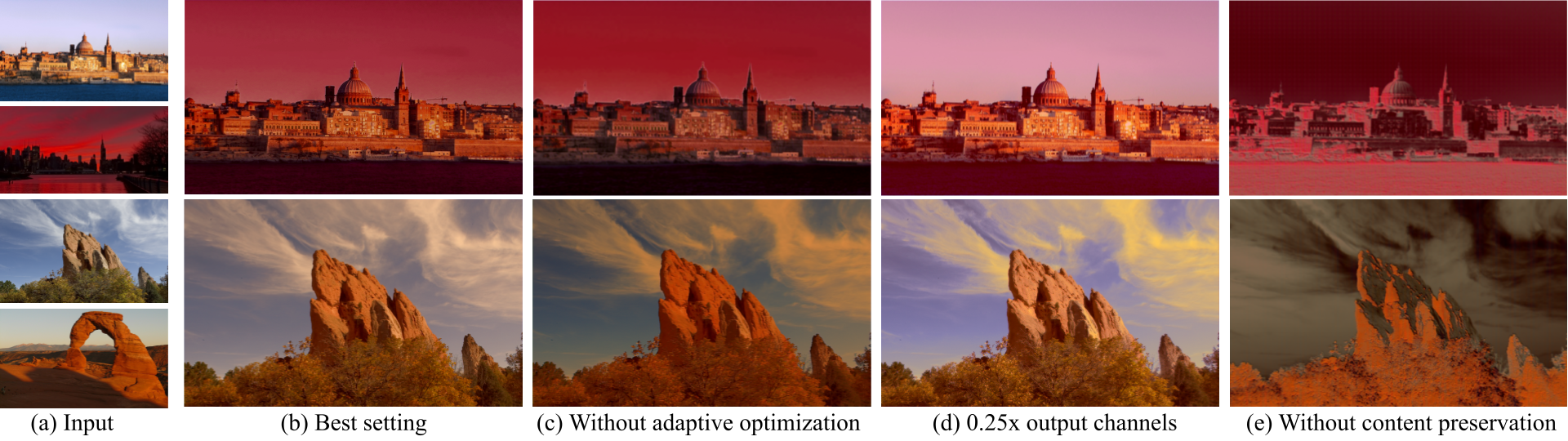}
    \caption{\textbf{Qualitative Comparison of Ablation Studies.}}
    \label{fig:ablation}
\end{figure*}

\subsection{Ablation Studies}
The comprehensive evaluation of ablation is depicted in \Cref{fig:ablation}, with (a) representing the content-style input pair and (b) denoting the optimal configuration introduced in \Cref{sec:approach}. The quantitative comparison table is provided as \Cref{tab:ablation_table} for reference.

\noindent\textbf{Instance-Adaptive Optimization.}
Instance-adaptive optimization serves to prioritize the photorealism of the result and accelerate the convergence of StyleNet. For comparison analysis, we configured epoch to $150$ with early stopping disabled and $\alpha$ to $1$ to produce the results without the adaptive optimization. \Cref{fig:ablation} (c) showcases that the absence of adaptive optimization leads to overfitting the style image and generating unrealistic artifacts. Moreover, \Cref{tab:ablation_table} reveals that the results obtained without adaptive optimization exhibit a high style similarity due to their overfitting to the style image, which negatively impacts content similarity. This leads to a lower F-1 score compared to results obtained with adaptive optimization. Additionally, it's worth noting that StyleNet converges 1.76 times faster when adaptive optimization is employed.

\noindent\textbf{Model Size.} StyleNet employs a lightweight structure to ensure the transfer effiency. We explored the effects of enlarging or reducing the size of StyleNet. In general, larger models can learn more complex transformations, but they might overfit with limited data, while smaller ones may underfit the data and cannot perform the transfer task well.  We tested this by making the model four times bigger and then a quarter of its original size.
The larger model in \Cref{fig:ablation} (c) has a higher style similarity but a lower content similarity, indicating overfitting to the style image. In contrast, the smaller model in \Cref{fig:ablation} (d) underfits the style image. Despite having more parameters, the larger model processed faster due to adaptive optimization. The smaller model, with fewer parameters, has its training extended by adaptive optimization to complete its convergence.

\noindent\textbf{Content Preservation.} StyleNet adopts a content shortcut and proper padding setting to safeguard the content information during the transfer process. \Cref{fig:ablation} (e) displays the results without these content preservation techniques. Notably, the absence of a content shortcut leads to severe overfitting to the style image, while the lack of proper padding results in the emergence of grid pattern artifacts.

\section{Conclusion}
\label{sec:conclusion}

We have developed a lightweight and adaptable Universal Photorealistic Style Transfer (UPST) approach, facilitating style transfers for high resolutions and various input formats without pre-training on annotated datasets or imposing extra constraints. Our experimental findings demonstrate that our approach outperforms others in terms of photorealism and the efficiency of multi-frame transfers.

Nonetheless, UPST has its own limitations. One challenge is finding meaningful content-style pairs for transformation, as improper input combinations can lead to unrealistic results. Additionally, UPST lacks the ability to specify specific regions for transformation. Promising directions for future development involve integrating semantic maps into UPST, which enables UPST to transfer specific regions of interest, and replacing VGG with other lightweight CNN models that further expedite the optimization process.

\clearpage

{\small
\bibliographystyle{ieee_fullname}
\bibliography{egbib}
}
\end{document}